\pdfoutput=1
\documentclass{article}
\usepackage{arxiv}
\usepackage[utf8]{inputenc} 
\usepackage[T1]{fontenc}    
\usepackage{hyperref}       
\usepackage{url}   
\usepackage{booktabs}       
\usepackage{amsfonts}       
\usepackage{nicefrac}       
\usepackage{microtype}      
\usepackage{graphicx}
\usepackage{epstopdf}
\usepackage{amsmath}
\usepackage{mwe}
\usepackage{color,soul}
\usepackage{import}
\usepackage{physics}
\usepackage{subcaption}
\usepackage[labelfont=bf]{caption}
\usepackage{cleveref}
\usepackage{array,multirow}
\crefdefaultlabelformat{#2\textbf{#1}#3} 
\crefname{figure}{\textbf{Fig.}}{\textbf{Fig.}}
\Crefname{figure}{\textbf{Fig.}}{\textbf{Fig.}}
\crefformat{equation}{Equation~#2#1#3}
\Crefformat{equation}{Equation~#2#1#3}
\crefname{table}{\textbf{Table}}{\textbf{Table}}
\Crefname{table}{\textbf{Table}}{\textbf{Table}}

\title{A Biologically Plausible Learning Rule for \\Deep Learning in the Brain}

\author{
  Isabella~Pozzi
   \\
   Vision \& Cognition Group\\
 Netherlands Institute for Neuroscience\\
 Amsterdam, The Netherlands\\
  \texttt{i.pozzi@nin.knaw.nl} \\
   ~\And~
Sander M.~Boht\'e \\
  Machine Learning Group\\
  Centrum Wiskunde \& Informatica\\
Amsterdam, The Netherlands\\
  \texttt{s.m.bohte@cwi.nl} \\
     ~\And~
 Pieter R.~Roelfsema \\
  Vision \& Cognition Group\\
 Netherlands Institute for Neuroscience\\
 Amsterdam, The Netherlands\\
 \texttt{p.roelfsema@nin.knaw.nl} \\
}

\begin{document}
\maketitle

\begin{abstract}
Intelligence is our ability to learn appropriate responses to new stimuli and situations, and significant progress has been made in understanding how animals learn tasks by trial-and-error learning. The success of deep learning in end-to-end learning on a wide range of complex tasks is now fuelling the search for similar deep learning principles in the brain. 
While most work has focused on biologically plausible variants of error-backpropagation, learning in the brain seems to mostly adhere to a reinforcement learning paradigm, and while biologically plausible neural reinforcement learning has been proposed, these studies focused on shallow networks learning from compact and abstract sensory representations. Here, we demonstrate how these learning schemes generalize to deep networks with an arbitrary number of layers. The resulting reinforcement learning rule is equivalent to a particular form of error-backpropagation that trains one output unit at any time. 
We demonstrate the learning scheme on classical and hard image-classification benchmarks, namely MNIST, CIFAR10 and CIFAR100, cast as direct reward tasks, both for fully connected, convolutional and locally connected architectures. We show that our learning rule - Q-AGREL - performs comparably to supervised learning via error-backpropagation, with this type of trial-and-error reinforcement learning requiring only 1.5-2.5 times more epochs, even when classifying 100 different classes as in CIFAR100. Our results provide new insights into how deep learning may be implemented in the brain. 
\end{abstract}


\section{Introduction}
Among the learning rules for neural networks, reinforcement learning has the important virtue of occurring in animals and humans. Hence, reinforcement learning by artificial neural networks can be used as a model for learning in the brain \citep{bishop1995neural}. Indeed, previous theories have suggested how powerful reinforcement learning rules inspired by artificial neural networks could be implemented in the brain \citep{roelfsema2018control} and the methodology for shaping neural networks with rewards and punishments is an active area of research \citep{schmidhuber2011fast,friedrich2010learning,vasilaki2009spike,o2006making,huang2013assembling}.

Current deep artificial neural networks are typically trained with variants of the error-backpropagation rule, a method that adjusts synaptic weights in multilayer networks to reduce the errors in the mapping of inputs into the lower layer to outputs in the top layer. It does so by first computing the output error, which is the difference between the actual and desired activity levels of output units, and then determines how the strength of connections between successively lower layers should change to decrease this error using gradient descent \citep{rumelhart1986general}. 

Similarly to deep neural networks, the brain of humans and animals are composed of many layers between the sensory neurons that register the stimuli and the motor neurons that control the muscles. Hence it is tempting to speculate that the methods for deep learning that work so well for artificial neural networks also play a role in the brain \citep{marblestone2016toward,scholte2017visual}. A number of important challenges need to be solved, however, and some of them were elegantly expressed by Francis Crick who argued that the error-backpropagation rule is neurobiologically unrealistic \citep{crick1989recent}. The main question is: how can the synapses compute the error derivative based on information available locally? In more recent years, researchers have started to address this challenge by proposing ways in which learning rules that are equivalent to error-backpropagation might be implemented in the brain \citep{urbanczik2014learning,schiess2016somato,roelfsema2005attention,rombouts2015attention,brosch2015reinforcement,richards2019dendritic,scellier2019equivalence,amit2018biologically,sacramento2018dendritic}, most of which were reviewed in \citep{marblestone2016toward}. 
One of the main challenges remained to inform synapses at the lower network levels about the desired change in their strength, because the influence of changes in their strength on activity in the output layer is only indirect and depends on many intermediate synapses. In addition, most of the algorithms still focus on learning high-rank representations, while animals learning to select actions by trial-and-error is intrinsically low-rank. 

Here we will focus on a particular type of learning rule known as AGREL (attention-gated reinforcement learning) and AuGMEnT (attention-gated memory tagging) \citep{roelfsema2005attention,rombouts2015attention}, which provide us with a biologically plausible (in particular, low-rank learning) solution for the lower synapses-update challenge. 
These learning rules realized that in a reinforcement learning setting the synaptic error derivative can be split into two factors: a reward prediction error (RPE) which is positive if an action selected by the network is associated with more reward than expected or if the prospects of receiving reward increase while it is negative if the outcome of the selected action is disappointing. In the brain, the RPE is signaled by neuromodulatory systems that project diffusely to many synapses so that they can inform them about the RPE \citep{schultz2002getting}; the second factor is an attentional feedback signal that is known to propagate from the motor cortex to earlier processing levels in the brain \citep{roelfsema2018control,pooresmaeili2014simultaneous}. When a network chooses an action, this feedback signal is most pronounced for those neurons and synapses that can be held responsible for the selection of this action and hence for the resulting RPE.
These two factors jointly determine synaptic plasticity. As both factors are available at the synapses undergoing plasticity, it has been argued that learning schemes such as AGREL and AuGMEnT are indeed implemented in the brain \citep{roelfsema2018control}.
However, the previous AGREL and AuGMEnT models used networks with a single hidden layer, and modeled learning in tasks with only a handful input neurons. 

The present work has two goals. The first is to establish the relation between the biologically realistic learning rules and error-backpropagation for deep networks composed of multiple layers between the input and output layer in a reinforcement learning setting. Can the brain, with its many layers between input and output indeed solve the credit-assignment problem in a manner that is equivalent to deep learning? The second goal is to compare trial-and-error learning with biologically plausible learning rules to learning with error-backpropagation in more challenging problems. To this aim we investigated if and how the biologically learning rules cope with different datasets, namely MNIST, CIFAR10 and CIFAR100, trained as direct reward reinforcement learning tasks. 

\section{Biologically plausible deep reinforcement learning}
\begin{figure}[t]
    \centering
    \includegraphics[width = .85\linewidth]{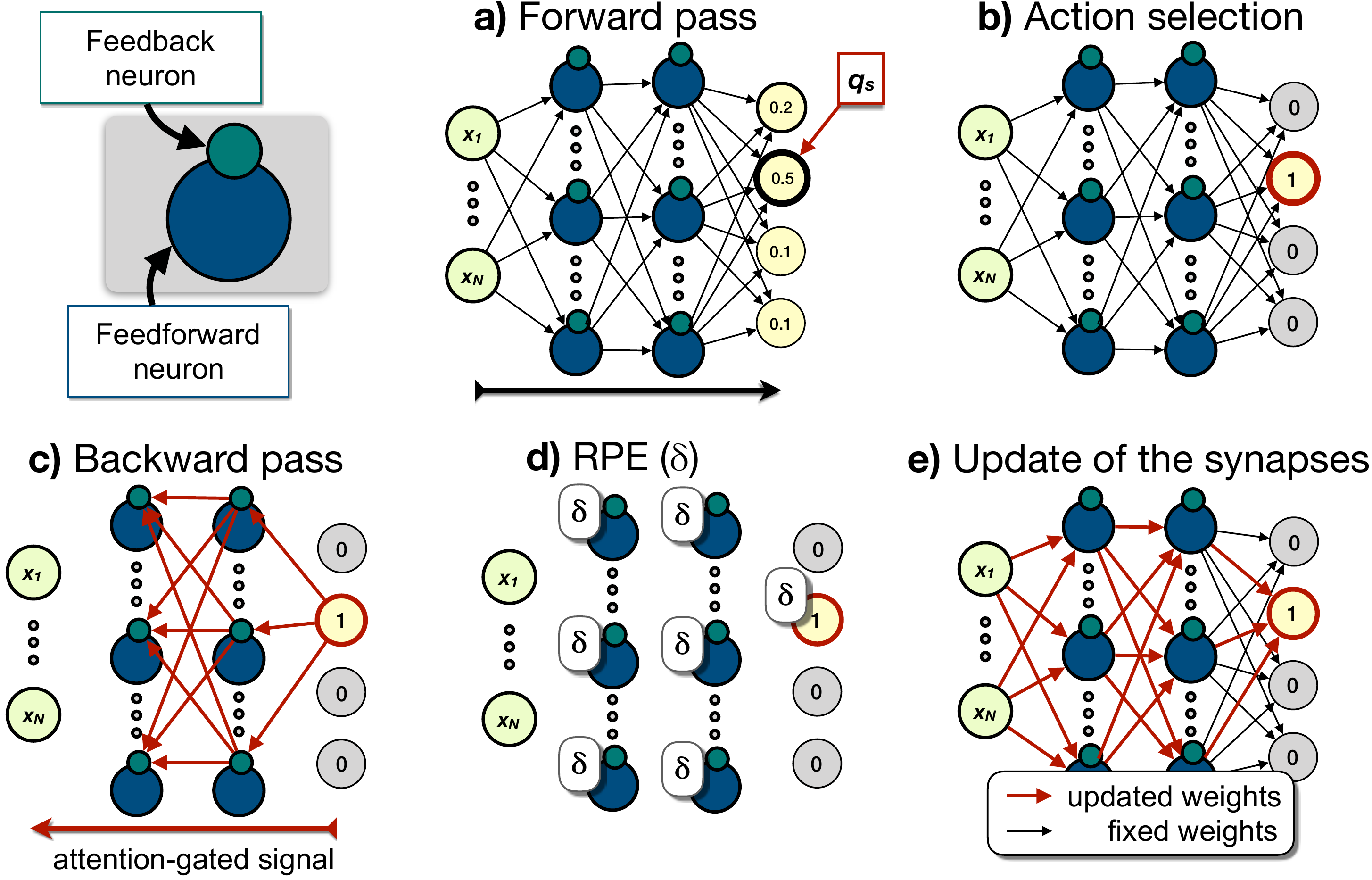}    \caption{Schematic depiction of Q-AGREL. At each node, a feedforward neuron (blue) and a feedback neuron (green) are present; separate feedforward and feedback weights connect the nodes in the network. }
    \label{fig:feedback}
\end{figure}

We here generalize and extend AGREL to networks with multiple layers with two modifications of the previous learning schemes. Firstly, we use rectified linear (ReLU) functions as activation function of the neurons in the network. This simplifies the learning rule, because the derivative of the ReLU is equal to zero for negative activation values, and has a constant positive value for positive activation values. Note however that this can easily be generalized to other activation functions. 
Secondly, we assume that network nodes correspond to cortical columns with feedforward and feedback subnetworks: in the present implementation we use a feedforward neuron and a feedback neuron per node, shown as blue and green circles in \cref{fig:feedback}.

Overall, the network learning goes through five phases upon presentation of an input image:
the signal is propagated through the network by feedforward connections to obtain activations for the output units where the Q-values are computed ({\bf Forward pass}, \cref{fig:feedback}\!~\textbf{a}), in the output layer one output unit wins in a stochastic, competitive action selection process ({\bf Action selection}, \cref{fig:feedback}\!~\textbf{b}), the selected output unit causes (attention-like) feedback to the feedback unit of each node ({\bf Backward pass}, \cref{fig:feedback}\!~\textbf{c}, note that this feedback network propagates information about the selected action, just as in the brain, see e.g. \cite{roelfsema2018control}, and that it does not need to propagate error signals, which would be biologically implausible). A reward prediction error $\delta$ is globally computed (\cref{fig:feedback}\!~\textbf{d}) after the outcome of the action is evident, and the strengths of the synapses (both feedforward and feedback) are updated (\cref{fig:feedback}\!~\textbf{e}). 

The proposed learning rule, Q-AGREL, has four factors:
\begin{equation}
    \Delta w_{i,j} = pre_i \cdot post_j \cdot \delta \cdot fb_j\,,
    \label{eq:1}
\end{equation}
where $\Delta w_{i,j}$ is the change in the strength of the synapse between units $i$ and $j$, $pre_i$ is a function of the activity of the presynaptic unit, $post_j$ a function of the activity of the postsynaptic unit and $fb_j$ the amount of feedback from the selected action arriving at feedback unit $j$ through the feedback network. This local learning rule governs the plasticity of both feedforward and feedback connections between the nodes.

The role of the feedback units in each node is to gate the plasticity of feedforward connections (as well as their own plasticity): $fb_j$ acts as a plasticity-gating term, which determines the plasticity of synapses onto the feedforward neuron. 
There is neuroscientific evidence for the gating of plasticity of feedforward connections by the activity of feedback connections, as was reviewed by \cite{roelfsema2018control}. 

\begin{figure}[t!]
  \begin{subfigure}[t!]{.355\linewidth}
  \includegraphics[width=.9\linewidth]{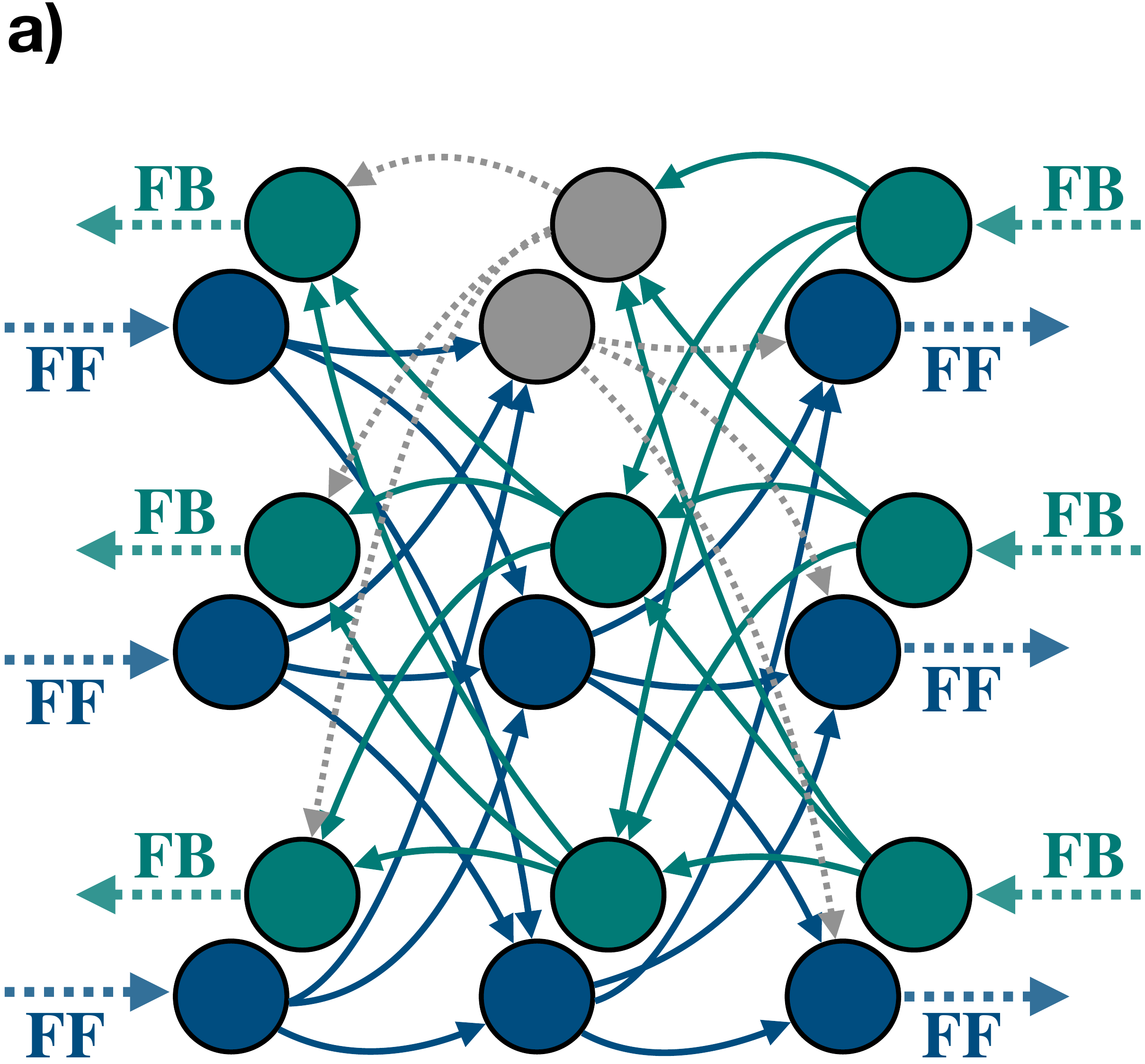}
  \end{subfigure}
  \hfill
  \begin{subfigure}[t!]{.3\linewidth}
  \includegraphics[width=.9\linewidth]{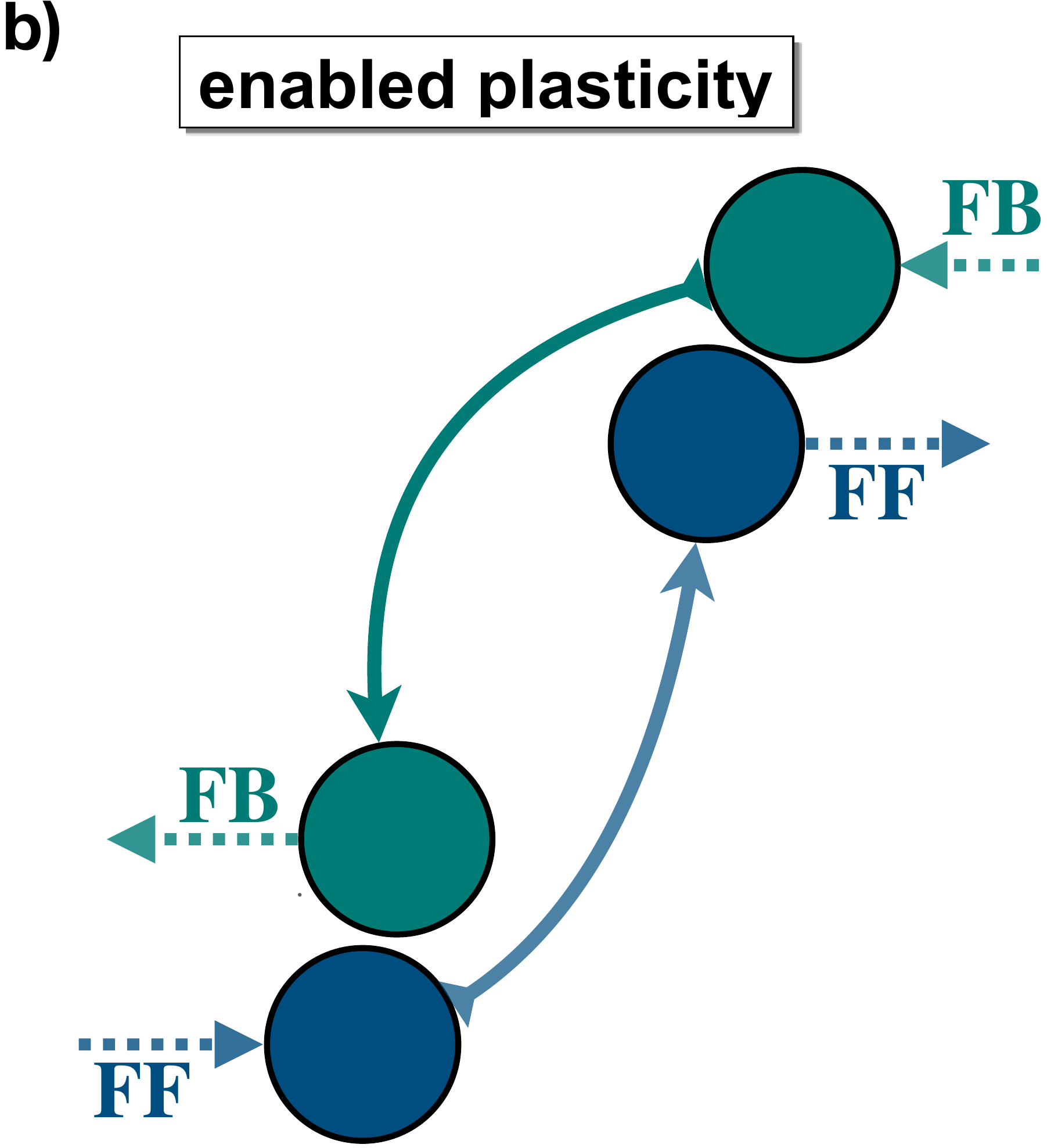}
  \end{subfigure}
  \hfill
  \begin{subfigure}[t!]{.3\linewidth}
  \includegraphics[width=.9\linewidth]{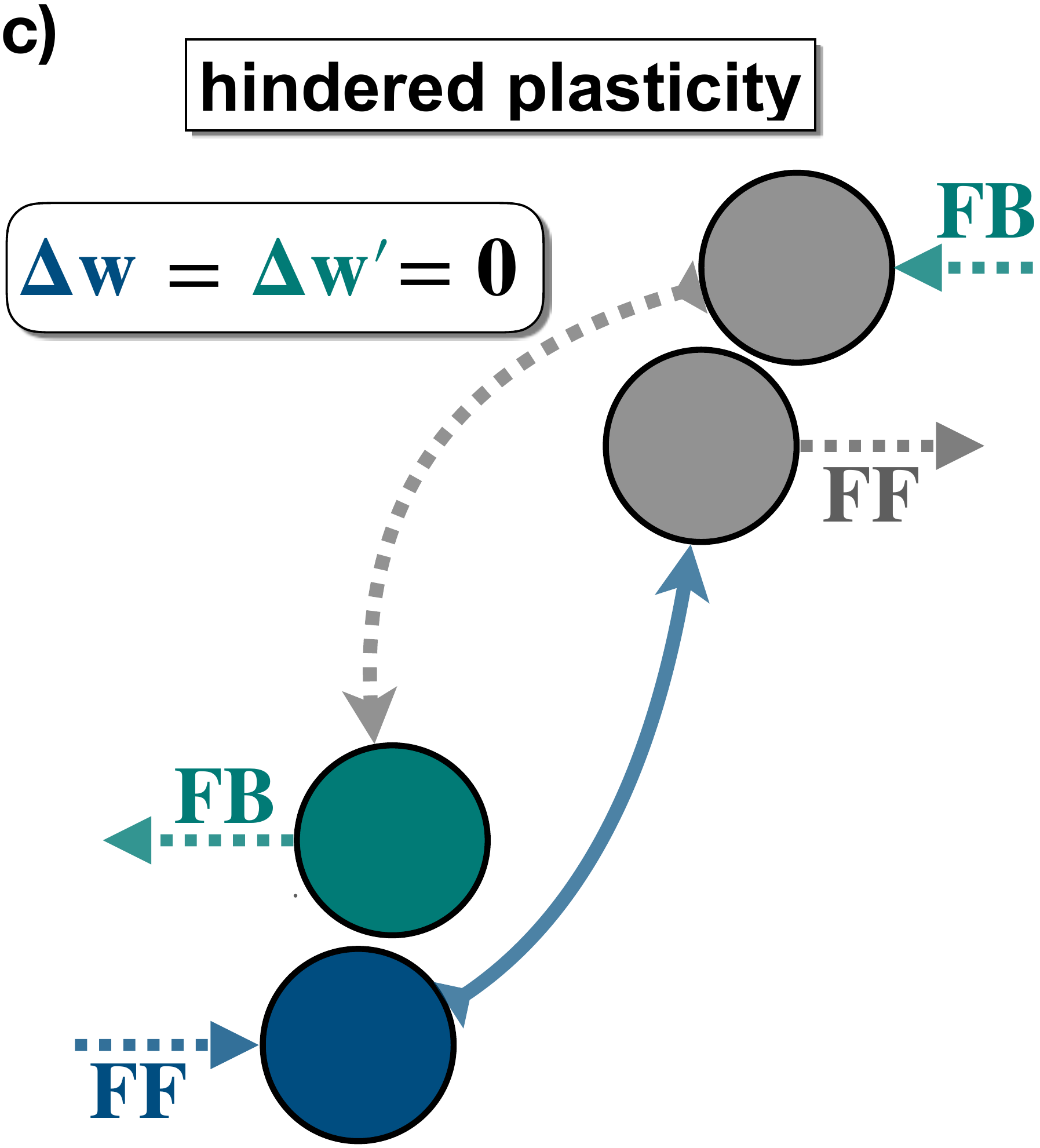}
  \end{subfigure}
      \caption{Q-AGREL algorithm plasticity gating. {\bf a)} Example hidden layers of a network; {\bf b)} when the activity of the feedforward neuron is above the threshold, the feedback signal is propagated to lower neurons and plasticity is enabled; {\bf c)} when the input to the feedforward unit stays below the threshold for activation the feedback signal is not propagated to the lower layer and plasticity is hindered.}
    \label{fig:agreldet}
\end{figure}

In the opposite direction, the feedforward units gate the activity of the feedback units. In \cref{fig:agreldet}\!~\textbf{a} examples of such interaction are shown. Feedback gating is shaped by the local derivative of the activation function $g_j$, which, for a unit with a ReLU activation function, corresponds to an all-or-nothing gating signal: for ReLU feedforward units, the associated feedback units of a node are only active if the feedforward units are activated above their threshold (\cref{fig:agreldet}\!~\textbf{b}), otherwise the feedback units remain silent and they do not propagate the feedback signal to lower processing levels (\cref{fig:agreldet}\!~\textbf{c}).
Gating of the activity of feedback units by the activity of feedforward units is also in accordance with neurobiological findings: attentional feedback effects on the firing rate of sensory neurons are pronounced if the neurons are well driven by a stimulus and much weaker if they are not \citep{van2017effects,roelfsema2006cortical,treue1999feature}.

In what follows we will first consider learning by a network with two fully connected hidden layers comprised of ReLU units (as in \cref{fig:feedback}), and we will then explain why the proposed learning scheme can train networks with an arbitrary number of layers in a manner that provides synaptic changes that are equivalent to a particular form of error-backpropagation.

In the network with two hidden layers, there are $N$ input units with activities $x_i$. The activation of the $J$ neurons in the first hidden layer, $y^{(1)}_j$, is given by:
\begin{equation}
y^{(1)}_j = \text{ReLU}\left(a^{(1)}_j\right) \qquad \text{with} \quad a^{(1)}_j = \sum^N_{i=1} u_{i,j}x_i \,,
\end{equation}
where $u_{i,j}$ is the synaptic weight between the $i$-th input neuron and the $j$-th neuron in the first hidden layer, and the ReLU function can be expressed as: $\text{ReLU}(x) = max(0,x)$.

Similarly, the activations of the $K$ neurons in the second hidden layer, $y^{(2)}_k$, are obtained as follows:
\begin{equation}
    y^{(2)}_k = \text{ReLU}\left(a^{(2)}_k\right) \qquad \text{with} \quad a^{(2)}_k = \sum^J_{j=1} v_{j,k}y^{(1)}_j \,,
\end{equation}
with $v_{j,k}$ as synaptic weight between the $j$-th neuron in the first hidden layer and the $k$-th neuron in the second hidden layer.
The $L$ neurons in the output layer are fully connected (by the synaptic weights $w_{k,l}$) to the second hidden layer and will compute a linearly weighted sum of their inputs:
\begin{equation}
    q_l = \sum^K_{k=1} w_{k,l}y^{(2)}_k \,,
\end{equation}
which we treat as Q-values as defined in Reinforcement Learning \citep{sutton1998reinforcement}, from which actions (or classifications) are selected by an action selection mechanism.

For the action-selection process, we implemented a max-Boltzmann controller \citep{wiering1997hq}: the network will select the output unit with the highest Q-value as the winning unit with probability $1 - \epsilon$, and otherwise it will probabilistically select an output unit using a Boltzmann distribution over the output activations:
\begin{equation}
    \text{P}(z_l = 1) = \frac{\exp q_l}{\sum_l \exp q_l}\,.
\end{equation}
After the competitive action selection process, the activity of the winning unit $s$ is set to one and the activity of the other units to zero, i.e. $z_{l=s} =1$ and $z_{l\neq s} =0$. The network then receives a scalar reward $r$ and a globally available RPE $\delta$ is computed as $\delta = r - q_s$,
where $q_s$ is the activity of the winning unit (see \cref{fig:feedback}), i.e. the current estimate of reward expectancy, which is indeed coded for in the brain, leading to a global prediction error $E = \frac{1}{2}\delta^2 $. In a classification task, we set the direct reward to 1 when the selected output unit corresponds to the correct class, and we set the reward to 0 otherwise.

Next, only the winning output unit starts propagating the feedback signal -- the other output units are silent. This feedback passes through the feedback connections with their own weights $w'$ to the feedback neurons in the next layer, where the feedback signal is gated by the local derivative of the activation function, and then further passed to the next layer of feedback neurons through weights $v'$, and so on. Hence only neurons that receive bottom-up input participate in propagating the feedback signal, in accordance with neuroscientific finding. We will demonstrate that this feedback scheme locally updates the synapses of the network in a manner equivalent to a particular form of error-backpropagation. 

Given the learning rate $\alpha$, the update of the feedforward weights $w_{k,s}$ between the last hidden layer and the output layer (but the same rule holds for the corresponding set of feedback weights, indicated as $w'_{s,k}$) is given by:
\begin{equation}\label{eq:var1}
     \Delta w_{k,s} = \alpha \delta y^{(2)}_k z_s = \Delta w'_{s,k}  \quad \text{and } \quad \Delta w_{k,l\neq s} = 0 = \Delta w'_{l\neq s,k} \,.
\end{equation}
   The feedforward and feedback weights $v$ and $v'$ between the first and second hidden layer change as follows:
\begin{align}
    \begin{split}\label{eq:var2}
    \Delta v_{j,k} &= \alpha \delta y^{(1)}_j  g_{(2)_k}  w_{s,k} z_s  
    = \alpha \delta y^{(1)}_j g_{(2)_k} fb_{y^{(2)}_k} = \Delta v'_{k,j}\,,
    \end{split} 
    \quad &\text{with} \quad g_{(2)_k} = 
    \begin{cases} 
    1 \quad\text{if} \quad  y^{(2)}_k > 0 \,, \\
    0 \quad \text{otherwise} \,,
    \end{cases} \\
    fb_{y^{(2)}_k} &= \sum_{l} g_{(O)_l} w'_{l,k} z_{l} 
    = w'_{s,k} z_s \,,  
\end{align}
   which is the feedback coming from the output layer. 
Finally, the weights $u$ between the inputs and the first hidden layer are adapted as:
\begin{align}
    \begin{split}\label{eq:var3}
    \Delta u_{i,j} &= \alpha \delta x_i g_{(1)_j} \sum_k v'_{k,j} g_{(2)_k} w'_{s,k} z_s \\
       &= \alpha \delta x_i g_{(1)_j} \sum_k v'_{k,j} g_{(2)_k} fb_{y^{(2)}_k} 
      = \alpha \delta x_i g_{(1)_j} fb_{y^{(1)}_j} \,,   
    \end{split}
    \, &\text{with} \quad g_{(1)_j} =
    \begin{cases}
    1 \quad\text{if} \quad  y^{(1)}_j > 0 \,, \\
    0 \quad \text{otherwise}\,,
    \end{cases}  \\
    fb_{y^{(1)}_j} & = \sum_k  g_{(2)_k} v'_{kj} fb_{y^{(2)}_k} \,,
    \end{align}
which is the feedback coming from the second hidden layer. 
$fb_{y^{(2)}_k}$ and $fb_{y^{(1)}_j}$ represent the activity of feedback neurons $y^{(2)}_k$ and $y^{(1)}_j$, which are activated by the propagation of signals through the feedback network once an action has been selected. 

In general, for deeper networks, updates of feedforward synapses $\Delta w _ {p,m}$ from $p$-th neuron in the $n$-th hidden layer onto $m$-th feedforward neuron in the $(n + 1)$-th hidden layer are thus computed as:
\begin{equation}
    \Delta w_{p,m} = \alpha \delta y^{(n)}_p g_{{(n+1)}_m} fb_{y^{(n+1)}_m} \label{eq:dwn} \,,
\end{equation}
and it is equal to the update of the corresponding feedback synapse $\Delta w'_{m,p}$, 
where the activity of the feedback unit is determined by the feedback signals coming from the $(n + 2)$-th hidden layer as follows:
\begin{equation}
    fb_{y^{(n+1)}_m}  = \sum_{q}  g_{{(n+2)}_q} v'_{q, m} fb_{y^{(n+2)}_q} \label{eq:fbn} \,,
\end{equation}
with $q$ indexing the units of the $(n + 2)$-th hidden layer.

The update of a synapse is thus expressed as the product of four factors: the RPE $\delta$, the activity of the presynaptic unit, the activity of postsynaptic feedforward unit and the activity of feedback unit of the same postsynaptic node, as anticipated in \cref{eq:1}, of which \cref{eq:var1,eq:var2,eq:var3} are all variants. Notably, all the information necessary for the synaptic update is available locally, at the synapse. Moreover, simple inspection shows tat the identical update for both feedforward and corresponding feedback synapses (i.e., $\Delta w_{k,l}$ and $\Delta w'_{l,k}$, $\Delta v_{j,k}$ and $\Delta v'_{k,j}$, and $\Delta u_{i,j}$ and $\Delta u'_{j,i}$) can be computed locally.

We next demonstrate that Q-AGREL is equivalent to a special form of error-backpropagation. In this form the network only computes the derivatives relative to the error of the Q-value of the selected output unit. 

For error-backpropagation in the same networks with error $E$ computed as the summed square error over all output Q-values $q_{l}$ and target outputs $\hat{q}_{l}$, $E = -\frac{1}{2}\sum_{l}({q}_{l}- \hat{q}_{l})^2$, if we define $\pdv{E}{q_l}=(q_l - \hat{q}_l) := e^{(O)}_l$, where the superscript $(O)$ stands for output layer, the relevant equations for the synaptic updates are:
\begin{align}
\Delta w_{k,l} & = -\alpha y^{(2)}_k e^{(O)}_l  \,, \\
\Delta v_{j,k} & = -\alpha y^{(1)}_j {y^{(2)}_k}' \underline{\sum_{l} w_{l,k} e^{(O)}_l}
= -\alpha y^{(1)}_j {y^{(2)}_k}' \underline{e^{(2)}_k}  \,, \\
\Delta u_{i,j} & = -\alpha x_i {y^{(1)}_j}' \underline{ \sum_{k} v_{k,j} {y^{(2)}_{k}}' \sum_{l} w_{l,k} e^{(O)}_l }
= -\alpha x_i {y^{(1)}_j}' \underline{ \sum_{k} v_{k,j} {y^{(2)}_{k}}' e^{(2)}_k }
= -\alpha  x_i {y^{(1)}_j}' \underline{e^{(1)}_j} \,, 
\end{align}
and in general, for a weight between units $p$ and $m$ in layer $n$ and $n+1$ respectively:
\begin{align}
\Delta w_{p,m} &= -\alpha y^{(n)}_{p} {y^{(n+1)}_m}' e_{m}^{(n+1)}\,, \qquad \text{with} \quad e_{m}^{(n+1)} =: \pdv{E}{y^{(n+1)}_m} = \sum_{q} w_{m,q} {y^{(n+2)}_q}' e^{(n+2)}_{q} \,,
\end{align}
with $q$ indexing the units of the $(n+ 2)$-th hidden layer and $y_{<\!\cdot\!>}^{(\cdot)'}$ 
indicating the derivative of $y^{(\cdot)}_{<\!\cdot\!>}$.
This corresponds to the Q-AGREL equations for the adjustment to the winning output when we set ${y^{()}_j}' = g_j$ and $e^{(O)}_{l=s} = \pdv{E}{q_s} = -\delta$, and $e^{(O)}_{l\neq s} = 0$:
\begin{align}
\Delta w_{k,l} & = \alpha \delta y^{(2)}_k  \,, \\
\Delta v_{j,k} & = \alpha \delta y^{(1)}_j g_k  fb_{y^{(2)}_k} \,, \\
\Delta u_{i,j} & = \alpha \delta  x_i g_j \sum_{k} g_{k} w_{l,k} fb_{y^{(2)}_k}  = \alpha \delta x_i g_j fb_{y^{(1)}_j}, \end{align}
and, by recursion, 
\begin{align}
\Delta w_{p,m} &=   \alpha \delta y^{(n)}_{p} g_m fb_{y^{(n+1)}_m}\,.
\end{align}
Compared to error-backpropagation, in the RL formulation of Q-AGREL only the error $e_l$ for the winning action $l$ is non-zero, and the weights in the network are adjusted to reduce the error for this action only. Depending on the action-selection mechanism, this trial-and-error approach will adjust the network towards selecting the correct action, while the \mbox{Q-values} for incorrect actions 
will only decrease in strength occasionally, when the stochastic action takes an explorative action, in contrast to standard error-backpropagation, which will continuously drive the values of the incorrect actions to the appropriate lower action values. Hence, reinforcement learning of Q-values by Q-AGREL is expected to be slower than learning with a fully supervised method such as error-backpropagation.
We will test these predictions in our simulations.

\section{Experiments}
We tested the performance of Q-AGREL on the MNIST, CIFAR10 and CIFAR100 datasets, which are classification tasks, and therefore simpler than more general reinforcement learning settings that necessitate the learning of a number of intermediate actions before a reward can be obtained. 
These types of tasks have been addressed elsewhere \citep{rombouts2015attention}.

The MNIST dataset consists of 60,000 training samples (i.e. images of 28 by 28 pixels), while the CIFAR datasets comprise 50,000 training samples (images of 32 by 32 by 3 pixels), of which 1,000 were randomly chosen for validation at the beginning of each experiment. 
We use a \emph{batch gradient} to speed up the learning process (but the learning scheme also works with learning after each trial, i.e. not in batches): 100 samples were given as an input, the gradients were calculated, divided by the batch size, and then the weights were updated, for each batch until the whole training dataset was processed (i.e. for 590 or 490 batches in total), indicating the end of an \emph{epoch}. At the end of each epoch, a validation accuracy was calculated on the validation dataset. An early stopping criterion was implemented: if for 20 consecutive times the validation accuracy had not increased, learning was stopped. 

We ran the same experiments with Q-AGREL and with error-backpropagation for neural networks with with three and four hidden layers. The first layer could be either convolutional or locally connected, the second layer was convolutional but with a stride of 2 in both dimensions, to which a dropout of 0.8 (i.e. 80\% of the neurons in the layer were silent) was applied, then either only one fully connected layer or two followed (with the last layer having a dropout rate of 0.3). At the level of the output layer (which had 10 neurons for MNIST and CIFAR10, while it was 10 times bigger for CIFAR100) for error-backpropagation a softmax was applied and a cross-entropy error function was calculated. We decided to test networks with locally connected layers because such an architecture could represent the biologically plausible implementation of convolutional layers in the brain (since shared weights are not plausible). Moreover, instead of using max pooling layers to reduce the dimensionality of the layer following the convolutions, we substituted such layers with convolutional layers with equal number of filters and kernel size, but with strides (2,2), as described in \citep{springenberg2014striving}. As argued by Hinton \citep{hinton2016can}, dropout is biologically plausible as well: by removing random hidden units in each training run, it simulates the regularisation process carried out in the brain by noisy neurons. 

In summary, we ran experiments with the following architectures: \\[.1cm]
a) \texttt{conv32 3x3; conv32 3x3 str(2,2); drop.8; (full 1,000;) full500; drop.3},\\
b) \texttt{loccon32 3x3; conv32 3x3 str(2,2); drop.8; (full 1,000;) full500; drop.3},\\[.1cm]
with 10 different seeds for synaptic weight initialization. All weights were randomly initialized within the range $[-0.02, 0.02]$ and the feedback synapses were identical to the feedforward synapses (strict reciprocity). For MNIST only we also performed a few experiments with fully connected networks, of which the weights were initialized in $[-0.05, 0.05]$. 

\section{Results}
\newcommand{\STAB}[1]{\begin{tabular}{@{}c@{}}#1\end{tabular}}
\begin{table}[t!]
  \caption{Results (averaged over 10 different seeds, the mean and standard deviation are indicated; in some cases - indicated with "*" -  only 9 out of 10 seeds converged).}
  \label{tab:table1}
  \centering
  \begin{tabular}{cllllrl}
    \toprule
&Rule & $1^{st}$ layer & Hidden units & $\alpha$ & Epochs [\#] & Accuracy [\%] \\ 
\midrule
    \multirow{10}{*}{\STAB{\rotatebox[origin=c]{90}{\textbf{MNIST}}}}
&Q-AGREL & Full & 1500-1000-500 & 5e-01 & 130 (54) & 98.33 (0.09) \\
&Error-BP & Full & 1500-1000-500 & 1e-01 & 92 (11) & 98.32 (0.04) \\
&Q-AGREL & Conv & 21632-5408-500 &  1e+00 & 44 (10) & 99.17 (0.05) \\ 
&Q-AGREL & Conv & 21632-5408-1000-500 &  1e+00 & 59 (21) & 99.16 (0.16) \\ 
&Error-BP & Conv & 21632-5408-500 &  1e-02 & 26 (12) & 99.19 (0.10) \\ 
&Error-BP & Conv & 21632-5408-1000-500 &  1e-02 & 33 (12) & 99.21 (0.17) \\ 
&Q-AGREL & LocCon & 21632-5408-500 &  1e+00 & 83 (13) & 99.04 (0.14) \\ 
&Q-AGREL & LocCon & 21632-5408-1000-500 &  1e+00 & 66 (15) & 98.30 (0.22) \\ 
&Error-BP & LocCon & 21632-5408-500 &  1e-02 & 31 (10) & 98.82 (0.20)\\ 
&Error-BP & LocCon & 21632-5408-1000-500 &  1e-02 & 24 (13) & 98.73 (0.41) \\ 
\midrule
\multirow{8}{*}{\STAB{\rotatebox[origin=c]{90}{\textbf{CIFAR10}}}}
&Q-AGREL & Conv & 28800-7200-500 &  1e+00 & 119 (38) & 72.70 (1.93) \\ 
&Q-AGREL & Conv & 28800-7200-1000-500 &  1e+00 & 115 (23) & 73.54 (1.35) \\ 
&Error-BP & Conv & 28800-7200-500 &  1e-03 & 116 (28) & 72.25 (1.30) \\ 
&Error-BP & Conv & 28800-7200-1000-500 &  1e-03 & 83 (21) & 71.25 (1.08) \\ 
&Q-AGREL & LocCon & 28800-7200-500 & 1e+00 & 160 (52) & 62.49 (2.46) \\ 
&Q-AGREL & LocCon & 28800-7200-1000-500 & 1e+00 & 173 (36) & 64.37 (2.41) \\ 
&Error-BP & LocCon & 28800-7200-500 & 1e-03 & 164 (32) & 64.35 (1.90) \\ 
&Error-BP & LocCon & 28800-7200-1000-500 & 1e-03 & 145 (16) & 64.65 (1.16)  \\
\midrule
\multirow{8}{*}{\STAB{\rotatebox[origin=c]{90}{\textbf{CIFAR100}}}}
&Q-AGREL & Conv & 28800-7200-500 &  1e+00 & 200 (32) & 34.93 (1.37)* \\
&Q-AGREL & Conv & 28800-7200-1000-500  & 1e+00 & 230 (30) & 34.90 (1.49)* \\ 
&Error-BP & Conv & 28800-7200-500  & 1e-03 & 113 (28) & 39.48 (1.04) \\ 
&Error-BP & Conv & 28800-7200-1000-500  & 1e-03 & 104 (24) & 36.79 (1.78) \\
&Q-AGREL & LocCon & 28800-7200-500 & 1e+00 & 310 (36) & 28.24 (1.12) \\ 
&Q-AGREL & LocCon & 28800-7200-1000-500 & 1e+00 & 343 (68)  & 29.39 (2.38) \\
&Error-BP & LocCon & 28800-7200-500 & 1e-03 & 176 (27) & 32.61 (1.29) \\ 
&Error-BP & LocCon & 28800-7200-1000-500 & 1e-03 & 156 (13) & 32.73 (0.78) \\
    \bottomrule
  \end{tabular}
\end{table}

\Cref{tab:table1} presents the results of simulations with the different learning rules. 
We trained networks with only three hidden layers and networks four hidden layers; these networks had an extra hidden layer with 1000 units. We used 10 seeds for each network architecture and report the results as \emph{mean (standard deviation)}. 

Our first result is that Q-AGREL reaches a relatively high classification accuracy of 99.17\% on the MNIST task, obtaining essentially the same performance as standard error-backpropagation both with the architectures with convolutions and straightforward fully connected networks. The convergence rate of Q-AGREL was a factor of 1.5 to 2 slower than that of error-backpropagation for networks with convolutional layers, while it was a factor of 2.5 slower in networks for locally connected layers, but performing slightly better than error-backpropagation. 

The results obtained from networks trained on the CIFAR10 dataset show that networks trained with Q-AGREL reached the same accuracy (if not higher) than with error-backpropagation. Additionally, the number of epochs required for the networks to meet the convergence criterion was also comparable. 

\Cref{tab:table1} also shows the results obtained from networks trained on CIFAR100. The final accuracy obtained with Q-AGREL was somewhat lower than with error-backpropagation. However, we still see that Q-AGREL is able to learn the CIFAR100 classification task with a convergence rate only 2 to 2.5 times slower than error-backpropagation and the rate for CIFAR10. These results shows that such trial-and-error learning rule can scale up to a 10 times higher number of classes with a penalty relatively small. 

\begin{figure}[t]
    \centering
    \begin{subfigure}[b]{.50\linewidth}
    \includegraphics[height=.6\linewidth]{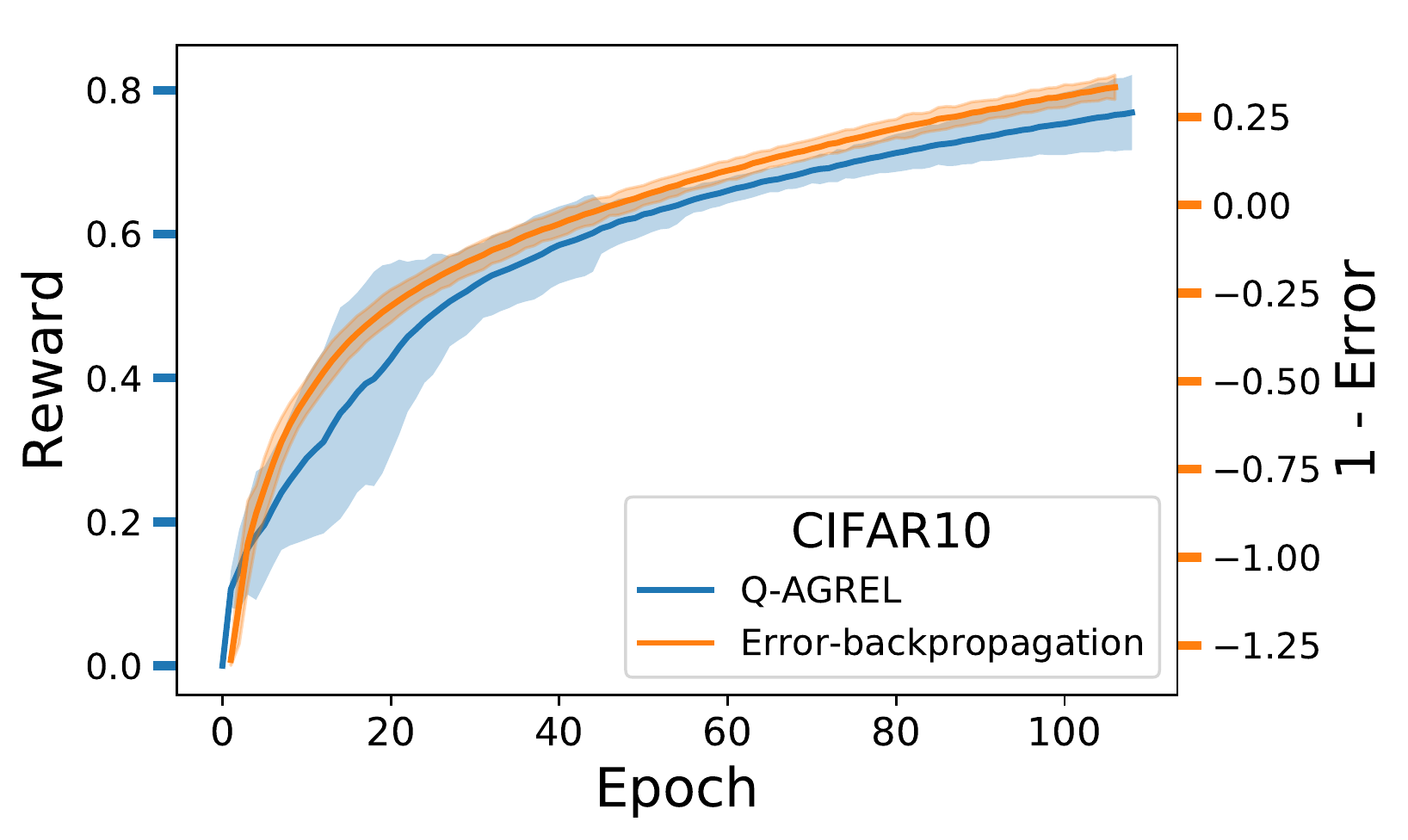}
    \end{subfigure}~\hfill\begin{subfigure}[b]{.50\linewidth}
    \includegraphics[height=.6\linewidth]{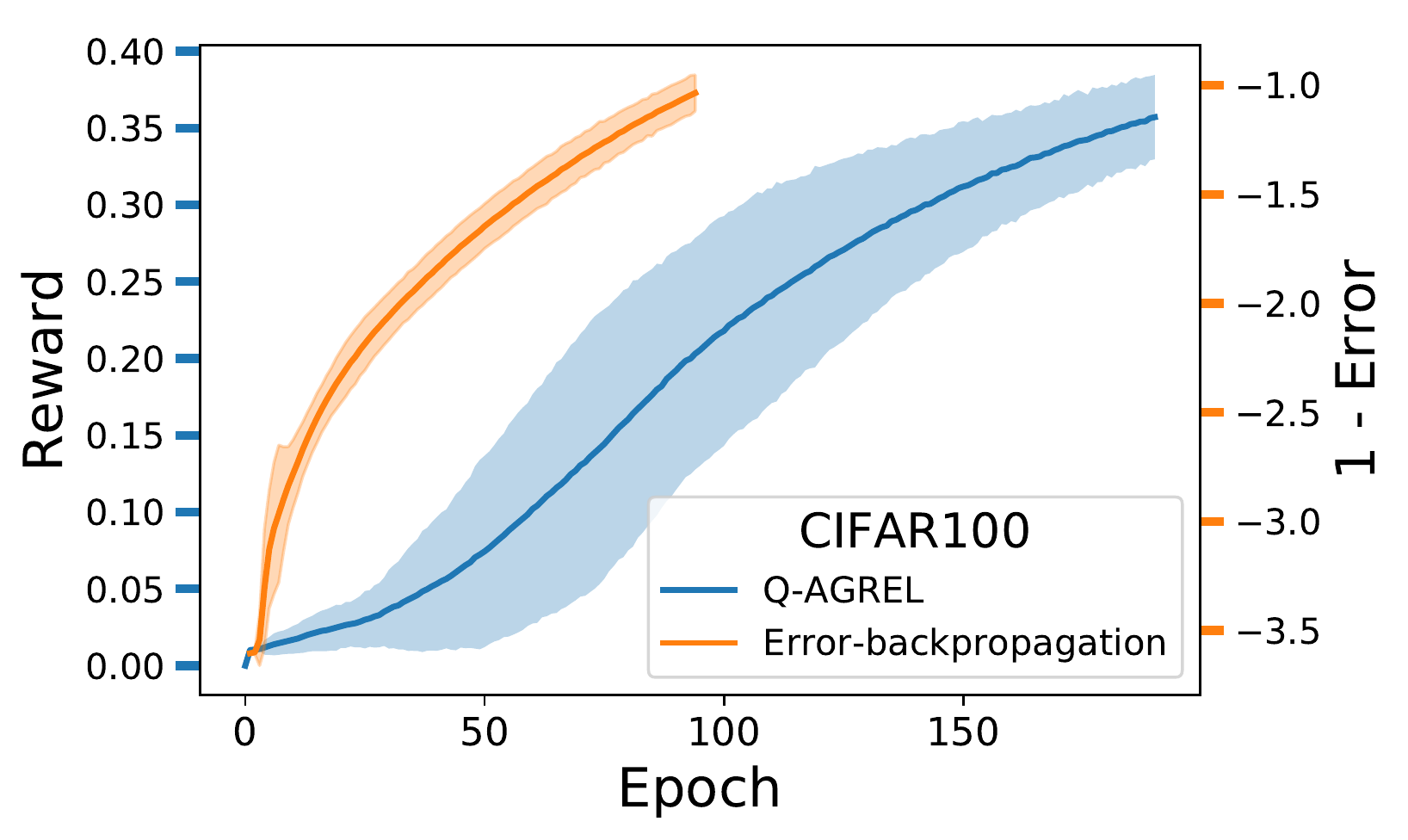}
    \end{subfigure}
    \caption{Example of learning process with the two learning schemes for CIFAR10 and CIFAR100.}
    \label{fig:rewer}
\end{figure}

To illustrate the learning process of networks trained with the Q-AGREL reinforcement learning approach, we show how the reward probability increases (\cref{fig:rewer}) during the training, compared to how the error (plotted as 1 - error) evolves throughout the epochs for 10 networks trained with error-backpropagation (in both cases, $mean \pm 2\sigma$ from 10 example networks is plotted as a function of the epochs), both for CIFAR10 (left panel) and CIFAR100 (right panel).
Indeed, learning with supervision is faster than learning by trial-and-error.

\section{Discussion}
We implemented a deep, biologically plausible reinforcement learning scheme called Q-AGREL and found that it was able to train networks to perform the MNIST, CIFAR10 and CIFAR100 tasks as direct reward problems with performance that was nearly identical to error-backpropagation. We also found that the trial-and-error nature of learning to classify with reinforcement learning incurred a very limited cost of 1-2.5x more training epochs to achieve the stopping criterion, even for classifying objects in 100 classes. 

The results were obtained with relatively simple network architectures (i.e. not very deep) and learning rules (no optimizers or data augmentation methods were used). These additions would almost certainly further increase the final accuracy of the Q-AGREL learning scheme. 

The present results demonstrate how deep learning can be implemented in a biologically plausible fashion in deeper networks and for tasks of higher complexity by using the combination of a global RPE and "attentional" feedback from the response selection stage to influence synaptic plasticity. Importantly, both factors are available locally, at many, if not all, relevant synapses in the brain \citep{roelfsema2018control}. We demonstrated that Q-AGREL is equivalent to a version of error-backpropagation that only updates the value of the selected action. Q-AGREL was developed for feedforward networks and for classification tasks where feedback about the response is given immediately after the action is selected. However, the learning scheme is a straightforward generalization of the AuGMeNT framework \citep{rombouts2012neurally,rombouts2015attention}, which also deals with reinforcement learning problems for which a number of actions have to be taken before a reward is obtained. 

We find it encouraging that insights into the rules that govern plasticity in the brain are compatible with some of the more powerful methods for deep learning in artificial neural networks. These results hold promise for a genuine understanding of learning in the brain, with its many processing stages between sensory neurons and the motor neurons that ultimately control behavior.

\paragraph{Acknowledgments:}
The work was supported by the Nederlandse Organisatie voor Wetenschappelijk Onderzoek (Natural Artificial Intelligence grant 656.000.002) and the European Union Seventh Framework Program (grant agreement 7202070 “Human Brain Project,” and European Research Council grant agreement 339490 “Cortic\_al\_gorithms”) awarded to P.R.R.

\bibliographystyle{apalike}
\bibliography{ms}

\end{document}